\documentclass[11pt]{article}
\usepackage{geometry}
\usepackage{amsmath,amssymb,amsfonts,bm}
\usepackage{amsthm}
\usepackage{graphicx}
\usepackage{mathtools}
\usepackage{xcolor}
\usepackage[linesnumbered, ruled]{algorithm2e}
\usepackage{cite}
\usepackage{booktabs}
\usepackage{multirow}
\usepackage[hidelinks,bookmarks=false]{hyperref}
\hypersetup{pdfauthor={}}
\usepackage{cleveref}

\newtheorem{remark}{Remark}

\newtheorem{example}{Example}

\newcommand{\xpost}{\bar{x}}
\newcommand{\xprior}{\bar{x}^-}
\newcommand{\Xpostbf}{\Sigma_{x,t-1}}

\newcommand{\Xpost}{\Sigma_{x,t}}

\newcommand{\Xprior}{\Sigma_{x,t}^-}
\newcommand{\Xprioropt}{\Sigma_{x,t}^{-,*}}

\newcommand{\Xpriornom}{\hat{\Sigma}_{x,t}^-}

\newcommand{\Xpostprev}{\Sigma_{x,t-1}}
\newcommand{\xnom}{\hat{x}_0^-}
\newcommand{\Xnom}{\hat{\Sigma}_{x,0}^-}

\newcommand{\Qdist}{\mathbb{Q}}
\newcommand{\Qhat}{\hat{\mathbb{Q}}}

\newcommand{\Gauss}{\mathcal{N}}

\newcommand{\Bures}{\mathcal{B}}

\newcommand{\real}[1]{\mathbb{R}^{#1}}
\newcommand{\symm}[1]{\mathbb{S}^{#1}}

\newcommand{\pd}[1]{\symm{#1}_{++}}

\DeclareMathOperator{\Tr}{Tr}
\DeclareMathOperator{\Cov}{Cov}

\Crefname{assumption}{Assumption}{Assumptions}
\Crefname{algorithm}{Algorithm}{Algorithms}
\Crefname{example}{Example}{Examples}
\Crefname{figure}{Fig.}{Figs.}

\title{WRAP: Wasserstein-Robust Adaptive Plug-in for\\ Robot Localization%
\thanks{This work was supported in part by the Information and Communications Technology Planning and Evaluation (IITP) grants funded by MSIT No. 2022-0-00124, No. 2022-0-00480 and No. RS-2021-II211343,
Artificial Intelligence Graduate School Program (Seoul National University), the National Research Foundation of Korea (NRF) grant funded by MSIT No. RS-2026-25477173 and No. RS-2026-25504174, the National Aeronautics and Space Administration (NASA) under Grant 80NSSC22M0070,
the National Science Foundation (NSF) under Grants CMMI 2135925, CPS 2311085 and IIS 2331878, and the Higher Education and Science Committee of the RA under Grant 24IRF-2B002.}
} %
\author{
Minhyuk Jang, Astghik Hakobyan, Jungjin Lee, Naira Hovakimyan, and
Insoon Yang%
\thanks{M. Jang and N. Hovakimyan are with the Department of Mechanical Science and Engineering, Grainger College of Engineering, University of Illinois Urbana-Champaign, Urbana, IL, USA {\tt\small \{jang64, nhovakim\}@illinois.edu}. A. Hakobyan is with the Center for Scientific Innovation and Education and the National Polytechnic University of Armenia, Yerevan, Armenia {\tt\small astghik.hakobyan@csie.am}.
J. Lee and I. Yang are with the Department of Electrical and Computer Engineering and Automation and Systems Research Institute, Seoul National University, Seoul, South Korea {\tt\small \{jungbbal, insoonyang\}@snu.ac.kr}. }
}

\date{}

\begin{document}

\maketitle


\begin{abstract}
Robotic localization under changing sensing conditions can suffer from biased errors and miscalibrated covariances. 
We present WRAP, an adapter-agnostic Wasserstein-robust plug-in for nonlinear extended Kalman filter (EKF) and error-state Kalman filter (ESKF) stacks. 
A causal module supplies time-varying effective process and measurement statistics; a mean-preserving Wasserstein local update then computes least-favorable covariances and a robust gain without changing the propagation model, residual, or retraction. This separates mean adaptation from covariance robustification and uses distinct radii for propagation and sensing. 
On 18 UWB--IMU sequences held out from adapter training, adapter-only and WRAP reduce mean 3-D position RMSE by $19.8\%$ and $27.4\%$ relative to the nominal ESKF; an isotropic ablation reaches $19.5\%$, linking the incremental gain to directional process-covariance redistribution. An in-sample GNSS--INS study shows that mean adaptation provides most of the accuracy gain, while DR improves consistency and mitigates over-tightened classical covariance estimates. The robust solve takes $0.05$\,ms for UWB and $2.92$\,ms for GNSS on a Jetson Orin Nano.
\end{abstract}


\section{Introduction}
\label{sec:intro}

\begin{figure}[t]
    \centering
    \includegraphics[
        width=0.8\linewidth]{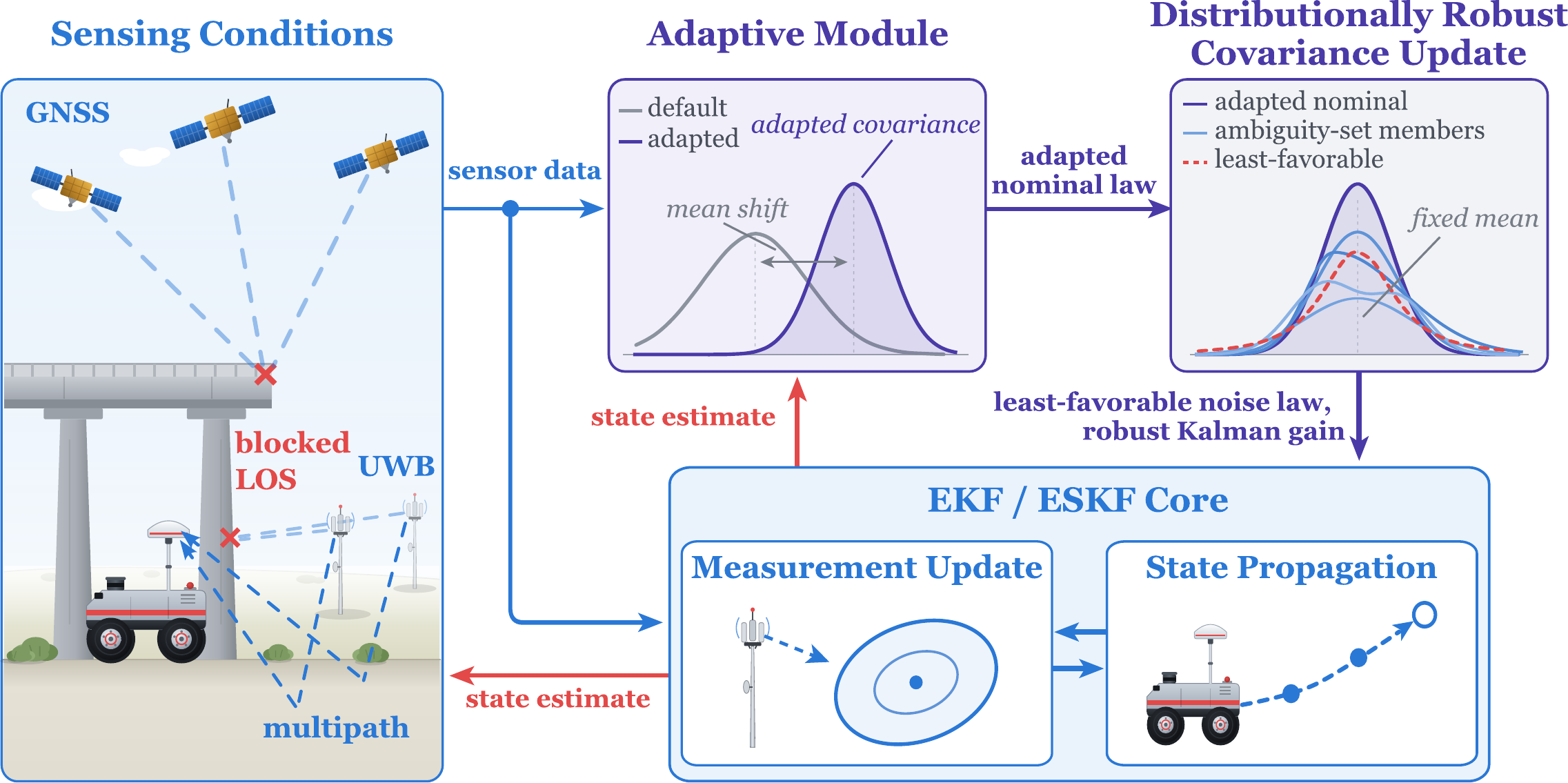}
    \caption{Overview of WRAP. The adaptive module supplies nominal process and measurement statistics. A Wasserstein DR local-estimation problem admits a covariance reformulation that yields least-favorable covariances and a robust Kalman gain, while the EKF/ESKF core remains unchanged.}
    \label{fig:diagram}
\end{figure}

Robot localization encounters abrupt sensing changes across outdoor and indoor operation. GNSS can be biased by multipath, UWB links can become non-line-of-sight (NLOS), and odometry, cameras, or LiDAR can degrade with contact or scene conditions~\cite{groves2013principles,zhao2024util,geneva2020openvins,xu2022fastlio2,hartley2020contact}. The resulting error law may change not only in magnitude but also in center, scale, correlation, or upper-quantile behavior, making a previously calibrated filter biased or miscalibrated; see~\Cref{fig:diagram}.

Extended Kalman filter (EKF) and error-state Kalman filter (ESKF) localization remain attractive because they combine nonlinear or manifold-valued propagation with closed-form local updates in real time~\cite{Kalman1960,sola2018micro}. Their performance, however, depends on the nominal process and measurement noise law supplied to each update. Nominal means enter the propagated state and predicted measurement, while nominal covariances determine the propagated and innovation
covariances and, consequently, the Kalman gain. Mean mismatch can therefore shift the innovation center and introduce systematic error, whereas covariance mismatch can misweight propagation and sensing. Because the two enter different parts of the recursion, correcting one does not resolve the other.

Particle, sigma-point, and iterated filters still require transition, likelihood, or noise models~\cite{gordon1993novel,julier2004unscented,bell1993iterated}. Adaptive Kalman filters estimate noise statistics through innovation matching, autocovariance, or Bayesian tracking~\cite{mehra1970identification,odelson2006new,sarkka2009recursive,huang2017novel,zhu2021adaptive,kruse2025adaptive,mohamed1999adaptive}, but their estimates can remain inaccurate under abrupt or unrepresented deployment shifts.
Learned filters and uncertainty models~\cite{revach2022kalmannet,kloss2021differentiable,liu2020tlio}
likewise supply gains or noise statistics from data, but remain nominal under
distribution shift.
 We ask how an existing real-time EKF/ESKF can adapt its nominal noise law while protecting its gain against residual covariance mismatch.

Distributionally robust (DR) estimation optimizes against the
worst-case distribution in an ambiguity set around a nominal
model~\cite{Wiesemann2014,mohajerin2018data,gao2023distributionally}.
Under Wasserstein ambiguity~\cite{villani2009optimal}, established
MMSE results for the affine--Gaussian, mean-preserving setting reduce
the minimax problem to least-favorable covariances and an affine
estimator~\cite{nguyen2023bridging}. Wasserstein DR Kalman filtering
has since been studied in several settings~\cite{si2023distributionally,
kargin2024distributionally,jang2025steady,jang2025distributionally},
with recent adaptive~\cite{wang2026stability} and residual-aware EKF
extensions~\cite{jang2026residual}. WRAP builds on this literature but
targets a different interface: whereas~\cite{jang2026residual} absorbs
linearization residuals into the ambiguity model, WRAP retains the
baseline EKF/ESKF linearization and couples separate mean-preserving
ambiguity sets on the process- and measurement-noise laws with any
causal nominal-law adapter. This division of labor is deliberate:
adaptation can update the nominal mean, whereas covariance
robustification cannot remove persistent mean error.

WRAP accepts effective process and measurement statistics from a learned, classical, Bayesian, or hybrid causal module, with unspecified components retained from the baseline. The local Wasserstein update robustifies the gain while leaving the nonlinear propagation, residual, gating structure, state representation, and retraction unchanged.

The contributions are:
\begin{itemize}
\item an adapter-agnostic noise-model interface for supplying any subset of effective process and measurement means and covariances to an existing EKF/ESKF;

\item a mean-preserving, noise-centric Wasserstein update with separate process and measurement ambiguity sets, using the established affine--Gaussian reduction without replacing model-specific filtering machinery; and

\item real-data component studies on 18 UWB--IMU sequences held out from adapter training, classical and isotropic ablations, an explicitly in-sample GNSS--INS mechanism study, and embedded runtime.
\end{itemize}


\section{Problem Formulation}
\label{sec:problem_formulation}
\subsection{Robot Model and Nominal Noise}
\label{subsec:robot_dynamics}
We consider a discrete-time nonlinear robot model
\begin{align*}
    x_{t+1} &= f_t(x_t,u_t,w_t),\\
    y_t &= h_t(x_t,v_t),
\end{align*}
where $x_t\in\mathcal X$ is the robot state, $u_t\in\real{n_u}$ is a known input, $y_t\in\real{n_y}$ is the measurement, and $w_t\in\real{n_w}$ and $v_t\in\real{n_v}$ are the process and measurement uncertainties. The state space $\mathcal X$ may be Euclidean or manifold-valued; $n_x$ denotes the dimension of the local coordinate in which the EKF/ESKF error and covariance are represented. The objective is recursive state estimation from the inputs and measurements available up to time $t$.

The formulation covers both conventional EKFs and error-state EKFs. When the state contains rotations or poses, the nominal state evolves on the corresponding manifold, while corrections are computed in local coordinates and injected through the baseline filter's retraction. WRAP operates only at this local filtering interface: it does not change the propagation model, residual definition, Jacobians, or state representation used by the baseline estimator.

\begin{example}[GNSS-aided inertial navigation]\label{ex:gnss}
Consider the GNSS--inertial navigation system (INS) in~\Cref{fig:diagram}, with state
\begin{equation*}
    x_t = \big(\, p_t,\; \dot{p}_t,\; R_t,\; b^a_t,\; b^g_t\,\big),
\end{equation*}
containing position, velocity, attitude, and IMU biases. 
Raw IMU readings drive propagation, while $w_t$ collects IMU noise, bias random walks, and residual propagation uncertainty. GNSS provides
\begin{equation}
    y_t = p_t + v_t,
    \label{eq:gnssins_meas}
\end{equation}
where $v_t$ is the GNSS position error. Near structures that block line of sight or cause multipath, $v_t$ may have both a nonzero mean and a miscalibrated covariance.
\end{example}

At the filtering interface, process and measurement uncertainty are represented by effective nominal Gaussian laws
\begin{equation}\label{eq:nominal_mod}
    \hat{\mathbb{Q}}_{w,t} = \mathcal{N}(\hat{w}_t,\hat{\Sigma}_{w,t}), \quad
    \hat{\mathbb{Q}}_{v,t} = \mathcal{N}(\hat{v}_t,\hat{\Sigma}_{v,t}),
\end{equation}
with $\hat{\Sigma}_{w,t}\in\pd{n_w}$ and $\hat{\Sigma}_{v,t}\in\pd{n_v}$. These nominal statistics may be fixed, sensor-reported, or produced by an adaptive module; they parameterize the local filtering surrogate and need not uniquely identify the physical sensor-noise law.
 In~\Cref{ex:gnss}, $(\hat v_t,\hat\Sigma_{v,t})$ describe the nominal GNSS position-error mean and covariance, while $(\hat w_t,\hat\Sigma_{w,t})$ describe the nominal IMU and propagation uncertainty. In particular, $\hat w_t$ may capture
residual systematic propagation error not represented by the
nominal bias states.

The nominal laws in~\eqref{eq:nominal_mod} are useful but
need not be exact. GNSS multipath, partial occlusion, and
degraded satellite geometry can bias the measurement error,
while imperfect inertial bias compensation, slip, or unmodeled
motion can corrupt the propagation model. The next subsection
makes the filtering interface explicit by identifying where
the nominal means and covariances enter the EKF/ESKF
recursion.

\subsection{Baseline EKF/ESKF Interface}
\label{subsec:local_model}

We summarize the nominal EKF/ESKF recursion to identify where the noise statistics enter. WRAP supplies adapted nominal statistics and replaces the nominal gain covariances with least-favorable ones.

Let $Y_t:=\{y_0,\ldots,y_t\}$ denote the measurement history. Before incorporating $y_t$, the filter has a prior estimate $\xprior_t$; after the update, it returns a posterior estimate $\xpost_t$. Let the corresponding estimation errors, expressed
in the baseline filter's local coordinates, be
\begin{equation*}
    e_t^- := x_t \boxminus \xprior_t,
    \qquad
    e_t := x_t \boxminus \xpost_t. 
\end{equation*}
For a Euclidean EKF, $\boxminus$ reduces to ordinary subtraction;
for an ESKF, it denotes the baseline local error map. In~\Cref{ex:gnss}, for example, the attitude error lies in the tangent space of $\mathrm{SO}(3)$ and is injected through the chosen retraction. We initialize the local prior law as $\Qhat_{x,0}^-=\Gauss(\xnom,\Xnom)$, with $\Xnom\in\pd{n_x}$.

Given the previous posterior estimate $\xpost_{t-1}$ and the nominal process noise mean $\hat w_{t-1}$, the prior state is
\begin{equation}
\xprior_t = f_{t-1}(\xpost_{t-1},u_{t-1},\hat w_{t-1}). \label{eq:prior_mean}
\end{equation}
The predicted measurement and innovation are
\begin{equation}
\hat y_t = h_t(\xprior_t,\hat v_t), \qquad \nu_t := y_t-\hat y_t.
\label{eq:predicted_measurement_innovation}
\end{equation}
Linearizing the propagation model around $(\xpost_{t-1},\hat w_{t-1})$ and the measurement model around $(\xprior_t,\hat v_t)$ gives the Jacobians $A_{t-1}:=\frac{\partial f_{t-1}}{\partial x},\quad G_{t-1}:=\frac{\partial f_{t-1}}{\partial w},\quad C_t:=\frac{\partial h_t}{\partial x},\quad D_t:=\frac{\partial h_t}{\partial v}$, each evaluated at its corresponding linearization points. In the GNSS--INS example, these are the standard INS error-state matrices~\cite{niu2025kfgins}; the position update~\eqref{eq:gnssins_meas} gives $C_t=[\,I\;\,0\;\,0\;\,0\;\,0\,]$ and $D_t=I$. We assume that \(D_t\) has full row rank. 
 
The resulting local prior-error and innovation models are
\begin{align}
    e_t^- &\approx A_{t-1}e_{t-1}+G_{t-1}(w_{t-1}-\hat w_{t-1}),
        \label{eq:local_prior_error}\\
    \nu_t &\approx C_t e_t^- + D_t(v_t-\hat v_t).
        \label{eq:local_innovation_model}
\end{align}
Using the nominal process covariance, the baseline prior covariance is
\begin{equation}
\Xpriornom = A_{t-1}\Xpostbf A_{t-1}^{\top} + G_{t-1}\hat\Sigma_{w,t-1}G_{t-1}^{\top}. 
\label{eq:nominal_prior_covariance}
\end{equation}
The nominal innovation covariance and Kalman gain are
\begin{equation}
    \hat S_t = C_t\Xpriornom C_t^\top + D_t\hat\Sigma_{v,t}D_t^\top,
    \quad
    \hat K_t = \Xpriornom C_t^\top \hat S_t^{-1}.
    \label{eq:nominal_gain}
\end{equation}
Finally, the posterior update is
\begin{equation}
\begin{aligned}
    \delta x_t=\hat K_t\nu_t,\qquad \xpost_t=\xprior_t\boxplus\delta x_t,\qquad
    \Xpost=\Xpriornom-\hat K_t\hat S_t\hat K_t^\top.
\end{aligned}
\label{eq:nominal_update}
\end{equation}
For a Euclidean EKF, $\boxplus$ denotes addition. For an ESKF, it denotes the baseline retraction, after which the baseline covariance-reset map is applied.

The recursion in~\eqref{eq:prior_mean}--\eqref{eq:nominal_update} exposes the two failure modes. The means $\hat w_{t-1}$ and $\hat v_t$ enter the propagated state and predicted measurement, so mean mismatch shifts the innovation center and can introduce systematic error. The covariances $\hat\Sigma_{w,t-1}$ and $\hat\Sigma_{v,t}$ determine the propagated and innovation covariances and, consequently, the Kalman gain and gating statistic. Covariance mismatch therefore
causes the filter to misweight propagation and sensing and may also alter measurement acceptance.

\section{Wasserstein-Robust Adaptive Plug-in}
\label{sec:dr_filter}

WRAP augments the noise-model interface of a baseline EKF/ESKF, as illustrated in~\Cref{fig:diagram}. A causal adaptive module supplies time-varying nominal process and measurement statistics. Centered on these estimates, a mean-preserving Wasserstein DR step selects least-favorable covariances for the local affine--Gaussian model and computes the robust Kalman gain. Adaptation therefore updates the nominal center of the noise law, while robustification protects the gain against residual covariance mismatch.

\subsection{Adaptive Module}
\label{subsec:adaptive}

The nominal law in~\eqref{eq:nominal_mod} need not be fixed. In WRAP, it is generated online by a causal adaptive module,
\begin{equation}
\begin{split}
    \mathcal A_t:
    (Y_{t-1},u_{0:t-1},s_t, \bar x_{t-1},\Sigma_{x,t-1})
    \mapsto
    (\hat w_{t-1},\hat\Sigma_{w,t-1},\hat v_t,\hat\Sigma_{v,t}),
\end{split}
    \label{eq:adaptive_map}
\end{equation}
where $s_t$ denotes current exogenous sensor metadata available
before forming the innovation, such as reported covariance,
signal quality, or measurement identity.

The module uses only information available before forming the
current innovation, including past measurements, inputs, filter
estimates, residuals, covariances, and available sensor metadata.
Neither the current measurement value $y_t$ nor its innovation
$\nu_t$ is used to predict the nominal law for the same update.
After the correction at time $t$, they may be appended to the
history used by $\mathcal A_{t+1}$.

The interface may use any causal classical, Bayesian, learned, or hybrid adaptive method~\cite{mehra1970identification,odelson2006new,
dunik2017noise,mohamed1999adaptive,sarkka2009recursive,
huang2017novel,zhu2021adaptive,kruse2025adaptive,
kim2026muse} and update any subset of the statistics in~\eqref{eq:adaptive_map}, with the remainder retained from the baseline. The resulting law is an effective local surrogate rather than a certified physical noise model: its means affect propagation and measurement prediction, while its covariances encode the confidence used by the filter.

\subsection{DR Local State Estimation via Covariance Update}
\label{subsec:dr}

The adaptive module supplies the nominal process and measurement laws
$\Qhat_{w,t-1}
=\mathcal N(\hat w_{t-1},\hat\Sigma_{w,t-1})$ and
$\Qhat_{v,t}
=\mathcal N(\hat v_t,\hat\Sigma_{v,t})$.
Because these laws may remain inaccurate, WRAP asks the following local minimax question: \emph{among nearby process and measurement laws, which ones make the current correction most difficult, and which correction minimizes the resulting worst-case error?}

We use the local models in~\eqref{eq:local_prior_error}--\eqref{eq:local_innovation_model} and, for a nominal law $\Qhat\in\mathcal P_2(\real{d})$ of a random vector $\xi$, define the mean-preserving Wasserstein ambiguity set
\begin{equation*}
\mathbb D(\Qhat,\theta)
:=
\left\{
\Qdist\in\mathcal P_2(\real{d})
\;\middle|\;
\begin{array}{l}
W_2(\Qdist,\Qhat)\leq\theta,\quad
\mathbb E_{\Qdist}[\xi]=\mathbb E_{\Qhat}[\xi]
\end{array}
\right\},
\label{eq:wass_ball}
\end{equation*}
where $\mathcal P_2(\real{d}):=\{\Qdist:\mathbb E_{\Qdist}\|\xi\|^2<\infty\}$ and $W_2$ is the type-2 Wasserstein distance~\cite{villani2009optimal}. The equality constraint preserves the adapted mean, while \(\theta\) quantifies residual distributional uncertainty around the nominal law.

Conditioned on $Y_{t-1}$, approximate the previous posterior error law by $\mathbb P_{e,t-1}:=\mathcal N(0,\Sigma_{x,t-1})$. For admissible $(\mathbb Q_{w,t-1},\mathbb Q_{v,t})$, define the product law $\mathbb P_t:=\mathbb P_{e,t-1}\otimes\mathbb Q_{w,t-1}\otimes\mathbb Q_{v,t}$.\footnote{As in standard local Kalman filtering, the previous estimation error, process uncertainty, and measurement uncertainty are assumed mutually independent. The derivation is exact for the affine--Gaussian stage and applied recursively as a local EKF/ESKF approximation.}
WRAP minimizes the worst-case first-order posterior mean-square error (MSE):
\begin{equation}
\min_{\psi_t (\cdot)}
\;\max_{\substack{
\mathbb Q_{w,t-1}\in
\mathbb D(\Qhat_{w,t-1},\theta_{w,t-1})\\
\mathbb Q_{v,t}\in
\mathbb D(\Qhat_{v,t},\theta_{v,t})}}
\mathbb E_{\mathbb P_t}
\!\left[\|e_t^- - \psi_t(\nu_t)\|^2\right].
\label{eq:minmax}
\end{equation}
Here, $\psi_t(\nu_t)$ is the local correction inferred from the innovation. The inner maximization stress-tests the update against admissible process and measurement laws, while the outer minimization selects the correction with the smallest worst-case MSE.

The two radii parameterize distinct sources of residual uncertainty. The process radius $\theta_{w,t-1}$ acts on propagation uncertainty, such as drift, slip, or unmodeled motion, while the measurement radius $\theta_{v,t}$ acts on sensing uncertainty, such as multipath, NLOS effects, or covariance miscalibration. Setting both radii to zero recovers the adapter-only update. Because $W_2$ is defined in the selected noise coordinates, the numerical radii inherit their units and scaling; they are therefore tuned and interpreted only within a fixed estimator parameterization.

\begin{remark}
The adaptive module sets the nominal means and covariances. The mean-preserving DR step keeps the adapted means fixed and robustifies only the covariances used in the Kalman gain. It can therefore change the relative weighting of propagation and sensing, but cannot correct an erroneous adapted mean.
\end{remark}

Under the local affine models, Gaussian nominal laws, mutual independence, and full row rank of $D_t$, established Wasserstein MMSE results imply an affine minimax correction and Gaussian least-favorable laws~\cite[Theorem~3.1 and Corollary~4.1]{nguyen2023bridging}. With $G_{t-1}$ and $D_t$ mapping the two noise terms, the noise-centric extension~\cite[Lemma~1 and Remark~3]{jang2025distributionally} reduces~\eqref{eq:minmax} to the covariance program below. This is a stagewise result for the linearized update, not a global minimax or stability result for the nonlinear recursion.

Because the adapted means are fixed, the Wasserstein constraints reduce to Bures--Wasserstein constraints~\cite{gelbrich1990formula,bhatia2019bures}. Following~\cite{nguyen2023bridging,jang2025distributionally}, redundant nominal eigenvalue lower bounds preserve the optimum while keeping the candidate covariances positive definite.
Given the previous posterior covariance $\Xpostprev$, the local minimax problem~\eqref{eq:minmax} admits the following equivalent covariance reformulation:
\begin{equation}
\begin{aligned}
\max_{\substack{\Sigma_{w,t-1},\Sigma_{v,t}\\ \Xprior, S_t}}
\quad &
\Tr\!\left(
\Xprior-\Xprior C_t^{\top}S_t^{-1}C_t\Xprior
\right)\\
\mathrm{s.t.}\quad
& \Bures^2(\Sigma_{w,t-1},\hat\Sigma_{w,t-1})
  \leq \theta_{w,t-1}^2\\
& \Bures^2(\Sigma_{v,t},\hat\Sigma_{v,t})
  \leq \theta_{v,t}^2\\
& \Xprior
=A_{t-1}\Xpostbf A_{t-1}^{\top}
  +G_{t-1}\Sigma_{w,t-1}G_{t-1}^{\top}\\
& S_t
=C_t\Xprior C_t^{\top}
  +D_t\Sigma_{v,t}D_t^{\top}\\
& \Sigma_{w,t-1}\succeq
  \lambda_{\min}(\hat\Sigma_{w,t-1})I_{n_w}\\
& \Sigma_{v,t}\succeq
  \lambda_{\min}(\hat\Sigma_{v,t})I_{n_v}.
\end{aligned}
\label{eq:robust_cov_problem}
\end{equation}
Here, $\Bures$ denotes the Bures--Wasserstein distance,\[
\Bures^2(\Sigma,\hat\Sigma)
:=
\Tr\left(
\Sigma+\hat\Sigma
-2(\hat\Sigma^{1/2}\Sigma\hat\Sigma^{1/2})^{1/2}
\right),
\qquad \Sigma,\hat\Sigma\succeq0.
\]

For fixed candidate covariances, the objective is the trace of the posterior error covariance $\Xpost$; the maximization therefore selects the covariances yielding the largest posterior MSE of the local affine--Gaussian surrogate.

Since $\hat\Sigma_{v,t}\succ0$, the measurement lower bound implies $\Sigma_{v,t}\succ0$. Together with the full row rank of $D_t$, this guarantees $S_t\succ0$, so its inverse is well defined.

Let $(\Sigma_{w,t-1}^*,\Sigma_{v,t}^*,\Xprioropt,S_t^*)$ maximize~\eqref{eq:robust_cov_problem}. The corresponding minimax correction in~\eqref{eq:minmax} has the affine Kalman form $\psi_t^*(\nu_t)=K_t^*\nu_t$, where
\begin{equation}
K_t^*=\Xprioropt C_t^{\top}(S_t^*)^{-1}.
\label{eq:robust_gain}
\end{equation}
The posterior update is
\begin{equation}
\begin{aligned}
\delta x_t^*=K_t^*\nu_t,\qquad
\xpost_t=\xprior_t\boxplus\delta x_t^*,\qquad
\Xpost=\Xprioropt-K_t^*S_t^*K_t^{*\top}.
\end{aligned}
\label{eq:robust_update}
\end{equation}
For an ESKF, the baseline error-state injection and covariance reset are applied after the correction.

Thus, the covariance reformulation implements the Wasserstein DR local-estimation problem in Kalman form. WRAP uses the adapted means in state propagation and measurement prediction, the least-favorable covariances in the gain computation, and the baseline state-injection and covariance-reset operations.


\subsection{Implementation and Algorithm}
\label{subsec:sdp}

WRAP approximately solves~\eqref{eq:robust_cov_problem} online using the warm-started Frank--Wolfe method of~\cite[Section~6]{nguyen2023bridging}. Since the objective in~\eqref{eq:robust_cov_problem} is concave in the decision covariances and the feasible set is convex and compact~\cite{nguyen2023bridging,gelbrich1990formula}, the Frank--Wolfe gap upper-bounds the suboptimality of the returned solution.
At each accepted update, the solver is warm-started from the preceding solution and returns least-favorable process and measurement covariances used to form the robust Kalman gain.

\Cref{alg:dr_adaptive_ekf} summarizes one WRAP update. The adaptive module first supplies the nominal noise law, after which the baseline filter computes the prior, innovation, and Jacobians. The baseline gate is evaluated using the adapted nominal innovation covariance \(\hat S_t\). If the measurement is accepted, the DR covariance step solves~\eqref{eq:robust_cov_problem}, forms the robust gain, and applies the baseline correction.

\begin{algorithm}[t]
\DontPrintSemicolon
\caption{WRAP EKF/ESKF update for \(t\geq1\)}
\label{alg:dr_adaptive_ekf}
\KwIn{Previous posterior \((\bar x_{t-1},\Sigma_{x,t-1})\), input \(u_{t-1}\), measurement \(y_t\), adaptive module \(\mathcal A_t\), radii \((\theta_{w,t-1},\theta_{v,t})\), gate radius \(\gamma\)}
Run \(\mathcal A_t\) to obtain \((\hat w_{t-1},\hat\Sigma_{w,t-1},\hat v_t,\hat\Sigma_{v,t})\)\;
Compute \(\bar x^-_t=f_{t-1}(\bar x_{t-1},u_{t-1},\hat w_{t-1})\) and \(\hat y_t=h_t(\bar x^-_t,\hat v_t)\)\;
Compute the innovation \(\nu_t=y_t-\hat y_t\)\;
Compute \(A_{t-1}\), \(G_{t-1}\), \(C_t\), \(D_t\), the nominal prior covariance \(\hat\Sigma_{x,t}^-\), and the nominal innovation covariance \(\hat S_t\) using~\eqref{eq:nominal_prior_covariance}--\eqref{eq:nominal_gain}\;

\eIf{\(\nu_t^\top\hat S_t^{-1}\nu_t>\gamma^2\)}{
    Set \(\bar x_t=\bar x_t^-\) and \(\Sigma_{x,t}=\hat\Sigma_{x,t}^-\)\;
}{
    Apply the Frank--Wolfe solver to~\eqref{eq:robust_cov_problem} to obtain \(\Sigma_{w,t-1}^*\) and \(\Sigma_{v,t}^*\), \(\Xprioropt\), and \(S_t^*\)\;
    Compute \(K_t^*\) using~\eqref{eq:robust_gain}\;
    Compute the local correction \(\delta x_t^*=K_t^*\nu_t\)\;
    Set \(\bar x_t=\bar x_t^-\boxplus\delta x_t^*\) and compute \(\Sigma_{x,t}\) using~\eqref{eq:robust_update}\;
    For an ESKF, apply the baseline covariance-reset map\;
}
Update the causal history available to \(\mathcal A_{t+1}\)\;
\end{algorithm}

\begin{remark}
The baseline gate precedes DR: rejected updates retain the adapted nominal prior, whereas accepted updates use the robust gain. Adaptation may change acceptance through $\nu_t$ and $\hat S_t$; WRAP does not claim robust gating or gross-outlier rejection.
\end{remark}
 
When both radii are zero, the ambiguity sets reduce to their adapted nominal laws, so the Frank--Wolfe solve can be skipped and the adapter-only EKF/ESKF update is recovered.


\section{Experimental Results}
\label{sec:experiments}

We evaluate WRAP on two robot-localization settings: UWB--IMU indoor localization and GNSS--INS outdoor localization.\footnote{Source code is available at \url{https://github.com/jangminhyuk/WRAP}}
Within each dataset, the baseline EKF/ESKF stack is held fixed. The variants differ only in the nominal noise law and, when the DR covariance step is enabled, the covariances used to form the Kalman gain. The experiments quantify trajectory root-mean-square error (RMSE) and examine when nominal-law
adaptation, covariance robustification, or their combination is most useful.

\subsection{Evaluation Protocol}
\label{subsec:evaluation_protocol}

\noindent{\bf Primary component comparison.}
Each sequence is evaluated using the same EKF/ESKF architecture under four variants:
\begin{enumerate}
    \item \textbf{Nominal}: fixed or sensor-provided nominal noise law, with \(\theta_w=\theta_v=0\).
    \item \textbf{Adapter-only}: adaptive nominal noise law, with \(\theta_w=\theta_v=0\).
    \item \textbf{DR-only}: fixed or sensor-provided nominal noise law with the DR covariance step enabled.
    \item \textbf{WRAP}: adaptive nominal noise law with the DR covariance step enabled.
\end{enumerate}
These variants isolate nominal-law adaptation, covariance robustification, and their combination.

\noindent{\bf Radius selection and data splits.}
The radii \(\theta_w\) and \(\theta_v\) are selected from $\{0,\allowbreak 0.01,\allowbreak 0.05,\allowbreak 0.1,\allowbreak 0.5,\allowbreak 1.0,\allowbreak 2.0,\allowbreak 5.0\}$. Because their values depend on the noise-coordinate parameterization and scaling, they are tuned separately for each estimator and are neither compared nor transferred across parameterizations.
For UWB--IMU, the adapter is trained on trials~1--3 and~7 and evaluated on trials~4--6. Each evaluation trial uses radii selected on the other two, so the trials are held out from adapter training but do not form a fully frozen hyperparameter test set. For GNSS--INS, the same six sequences are used for adapter training and evaluation; this is an in-sample mechanism study, not a generalization test. Its radii use leave-one-sequence-out selection. All variants share the metric, mask, and baseline estimator within each dataset.

\noindent{\bf Post-hoc diagnostics.}
Beyond trajectory RMSE, we use three ground-truth-based diagnostics to characterize the nominal error regime. For non-overlapping windows \(i\), let \(e_{i,k}\) denote the position-error samples and \(\bar e_i\) their mean. We define
\[
\begin{aligned}
\beta :=
\frac{\left(\sum_i\|\bar e_i\|^2\right)^{1/2}}
{\left(\sum_i\Tr\Cov_k(e_{i,k})\right)^{1/2}},\qquad
H_w :=
1-\frac{\operatorname{rms}\!\left(y_t-h_t(x_t^{\mathrm{gt}})\right)}
{\operatorname{rms}\!\left(h_t(\bar x_t^-)-h_t(x_t^{\mathrm{gt}})\right)}, \qquad
T := \frac{q_{0.99}(|r|)}{2.576}.
\end{aligned}
\]
Here, \(\beta\) compares windowwise bias with within-window scatter, with \(\beta>1\) indicating a bias-dominant regime. Process headroom \(H_w>0\) means that the measurement error is smaller than the prior-prediction error in measurement space, while \(T>1\) indicates upper-tail and/or scale mismatch relative to the scalar Gaussian reference. Residuals are mean-centered within each window and normalized per axis by the provider's nominal standard deviation. For UWB, \(r\) is the scalar standardized TDOA residual; for GNSS, it pools the standardized horizontal components. Windows are \(5\)\,s for UWB and \(30\)\,s for GNSS.

To summarize covariance-related headroom, we define the heuristic \emph{DR-headroom proxy} as \(\max\{H_w,T-1\}\). The diagnostics are computed per sequence and provider and summarized by provider medians. They are descriptive proxies, not theoretical conditions or calibrated predictors, and are never used for training, radius selection, gating, or online selection.

\noindent{\bf Scope of comparison.}
These are component studies within fixed EKF/ESKF stacks, not a comprehensive method ranking. Learned adapters use offline ground-truth supervision, whereas Sage--Husa and VB-AKF adapt online; the comparison tests interaction with WRAP, not equal-supervision architecture quality.

\subsection{Adaptive Modules and Implementation}
\label{subsec:adaptive_module_exp}

We evaluate two families of adaptive implementations for the interface in~\eqref{eq:adaptive_map}. The first consists of learned sequence adapters based on the multimodal architecture of MUSE~\cite{kim2026muse}. For UWB--IMU localization, we instantiate this architecture with Mamba~\cite{gu2023mamba}, Transformer~\cite{vaswani2017attention}, and gated recurrent unit (GRU)~\cite{cho2014gru} sequence backbones. GNSS--INS uses Mamba. The second family consists of two classical online adaptive estimators: Sage--Husa~\cite{sage1969adaptive} and a variational Bayesian adaptive Kalman filter (VB-AKF)~\cite{sarkka2009recursive}.

We adapt MUSE's multimodal architecture to predict nominal noise statistics rather than pose corrections. Its camera branch is replaced by historical UWB or GNSS measurements, the IMU branch uses a pretrained RoNIN encoder~\cite{herath2020ronin}, and the filter branch uses recent state estimates, residuals, and covariance features. For UWB--IMU, the output head predicts process and measurement noise means and covariances; for GNSS--INS, it predicts only the measurement-noise mean while retaining the receiver-reported covariance.

All learned-adapter inputs are causal: measurement-derived and residual features used at time $t$ end at $t-1$, so neither $y_t$ nor $\nu_t$ enters the current prediction. The UWB--IMU variants differ only in their sequence backbone and use width $d=128$, four blocks, and a 100-step window. The GNSS--INS model uses width $d=256$, two blocks, and a 10-step window. The nominal law is refreshed at 1\,Hz in both settings and retained between refreshes. These architectural choices are not required by WRAP, which only requires nominal statistics through \eqref{eq:adaptive_map}.\footnote{The DR covariance problem is solved using a warm-started Frank--Wolfe method, capped at 50 iterations with a duality-gap tolerance of $10^{-4}$.}

\subsection{UWB--IMU Localization}
\label{subsec:uwb_experiment}

\begin{figure}[t]
    \centering
    \includegraphics[width=0.75\linewidth]{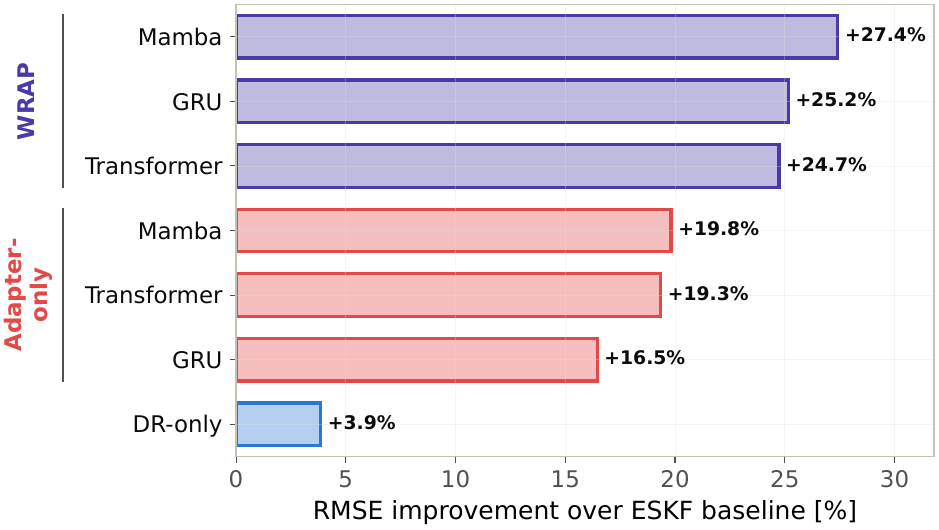}
    \caption{Mean 3-D position RMSE improvement on 18 UWB--IMU evaluation sequences (held out from adapter training; radii cross-validated).}
    \label{fig:uwb_improvement}
\end{figure}

We evaluate WRAP on 18 TDOA2 flight sequences from trials~4--6 of the UTIL dataset~\cite{zhao2024util}, all held out from adaptive-module training. The dataset provides UWB time-difference-of-arrival (TDOA) measurements, IMU data, and motion-capture reference trajectories from Crazyflie flights. We use its fixed-nominal ESKF baseline, with a scalar UWB TDOA residual at each update. Moving obstacles intermittently create NLOS conditions and time-varying measurement errors.

The learned adapters are trained end-to-end by unrolling the ESKF and minimizing position-tracking MSE. We also regularize the predicted measurement covariance with a Student-$t$ negative log-likelihood of the ground-truth measurement error at all measurement times, including gated-out updates. This provides heavy-tailed supervision for measurement-error scale beyond the tracking loss.

\Cref{fig:uwb_improvement} shows that adaptation provides most of the average improvement, while adding DR raises the observed average for all three backbones. This supports complementarity across the tested adapters but does not establish statistical superiority among backbones. Mamba has the highest observed average and is used below.

Because each UWB update is scalar, the full and isotropic measurement-side searches coincide. We therefore compare Mamba-based WRAP with an ablation that restricts both covariances to isotropic inflations of their nominal values, using the same radius-selection protocol.\footnote{For $c\geq1$, $\Bures(c\hat\Sigma,\hat\Sigma) =(\sqrt c-1)\sqrt{\Tr(\hat\Sigma)}$, so each radius determines one inflation factor $c(\theta)=\left(1+\theta/\sqrt{\Tr(\hat\Sigma)}\right)^2$.} The isotropic ablation achieves $19.5\%$ improvement, close to adapter-only at $19.8\%$, whereas the full covariance search achieves $27.4\%$. Within this ablation, the observed incremental gain is therefore attributable to process-side redistribution of uncertainty across directions rather than scalar inflation alone.

\Cref{tab:uwb_rmse} reports per-constellation RMSE for the Mamba backbone. WRAP achieves the lowest mean RMSE on constellations~\#1, \#2, and~\#4. Constellation~\#3 is the exception: the adapter is not beneficial in this setting, and DR-only performs best. This contrast motivates the post-hoc regime analysis below.

\begin{table}[t]
\centering
\caption{UWB--IMU TDOA2 localization RMSE (m) on the held-out trials, reported as mean\,$\pm$\,standard deviation across sequences in each anchor constellation.}
\label{tab:uwb_rmse}
\setlength{\tabcolsep}{4pt}
\footnotesize
\begin{tabular}{lcccc}
\toprule
Const. & ESKF & DR-only & Adapter-only (Mamba) & WRAP (Mamba)\\
\midrule
\#1 & {0.116\,$\pm$\,0.005} & {0.115\,$\pm$\,0.005} & {0.071\,$\pm$\,0.010} & {\textbf{0.065\,$\pm$\,0.009}} \\
\#2 & {0.107\,$\pm$\,0.011} & {0.106\,$\pm$\,0.010} & {0.069\,$\pm$\,0.006} & {\textbf{0.067\,$\pm$\,0.008}} \\
\#3 & {0.211\,$\pm$\,0.012} & {\textbf{0.178\,$\pm$\,0.018}} & {0.224\,$\pm$\,0.016} & {0.209\,$\pm$\,0.012} \\
\#4 & {0.487\,$\pm$\,0.108} & {0.487\,$\pm$\,0.108} & {0.375\,$\pm$\,0.075} & {\textbf{0.327\,$\pm$\,0.053}} \\
\bottomrule
\end{tabular}
\end{table}

\begin{figure}[t]
    \centering
    \includegraphics[width=\linewidth]{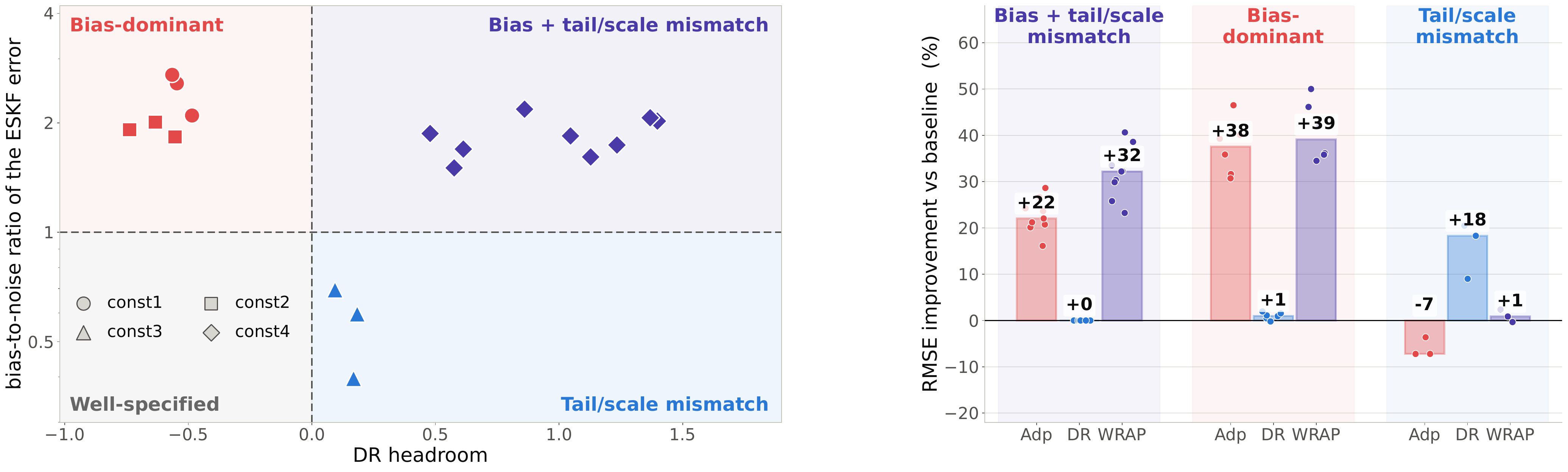}
    \caption{Error-regime analysis on the UTIL TDOA2 sequences, with Mamba used for the adaptive variants. \textbf{Left}: each sequence is positioned by the bias-to-noise ratio and DR headroom of the nominal ESKF; background shading denotes the diagnostic regimes, and marker shape denotes the anchor constellation. \textbf{Right}: median RMSE improvement over the nominal ESKF for adapter-only, DR-only, and WRAP in each populated regime; dots denote individual sequences.}
    \label{fig:uwb_regimes}
\end{figure}

For all UWB sequences, $H_w\leq0$, so the DR-headroom proxy is determined by $T-1$.
As shown in~\Cref{fig:uwb_regimes}, constellations~\#1 and~\#2 occupy the bias-dominant regime, where adapter-only and WRAP achieve median improvements of $38\%$ and $39\%$, respectively. Constellation~\#4 exhibits both bias and tail/scale mismatch; WRAP achieves $32\%$, compared with $22\%$ for adapter-only and $0\%$ for DR-only. Constellation
\#3 lies in the low-bias, tail/scale-mismatch regime, where adapter-only degrades RMSE by $7\%$ and DR-only achieves an $18\%$ median improvement. These post-hoc associations suggest a possible role for regime-aware selection, but the diagnostics require ground truth and are neither an online rule nor evidence of a causal performance predictor.

\begin{table}[t]
\centering
\caption{UWB--IMU TDOA2 mean RMSE improvement (\%) over the fixed-nominal ESKF (\(0.230\)\,m); DR headroom denotes the heuristic proxy.}
\label{tab:uwb_nominal}
\setlength{\tabcolsep}{3pt}
\footnotesize
\begin{tabular*}{0.8\linewidth}{@{\extracolsep{\fill}}lcrrr@{}}
\toprule
Nominal-law provider & \(\beta\) & DR headroom & Without DR & With DR \\
\midrule
Fixed           & 1.7 & $+0.31$ & $0.0$  & $3.9$ \\
Learned adapter & 1.1 & $+0.42$ & $19.8$ & $\mathbf{27.4}$ \\
Sage--Husa      & 1.7 & $+0.83$ & $-3.1$ & $1.8$ \\
VB-AKF          & 1.7 & $+0.66$ & $-2.5$ & $1.3$ \\
\bottomrule
\end{tabular*}
\end{table}

The adaptive interface is not restricted to learned models. VB-AKF adapts the measurement covariance from innovations, while Sage--Husa adapts both the measurement mean and covariance. As shown in \Cref{tab:uwb_nominal}, both perform below the fixed-nominal baseline when used without the DR step. One possible explanation is that a persistent measurement bias partly absorbed into the estimated state may leave only a weak signature in the subsequent innovation sequence, limiting innovation-driven mean and covariance adaptation. In contrast, the learned adapter receives supervision from ground-truth trajectory and measurement-error signals during training, which can expose systematic error patterns that are difficult to identify from online innovations alone.

The post-hoc diagnostics are consistent with this difference: the classical adapters leave $\beta$ at the baseline value of $1.7$, whereas the learned adapter reduces it to $1.1$. Applying the DR covariance step raises the classical variants slightly above the baseline, while WRAP with the learned adapter achieves the largest improvement, $27.4\%$.

\subsection{GNSS--INS Localization}
\label{subsec:gnss_ins_experiment}

We evaluate WRAP on the i2Nav-Robot dataset \cite{tang2025i2navrobot}, which provides robot trajectories with MEMS IMU, GNSS, and post-processed reference data. The implementation is based on KF-GINS~\cite{niu2025kfgins}, whose state includes position, velocity, attitude, and inertial biases. Each GNSS update provides a three-dimensional position measurement. 

The Mamba adapter is trained by supervised maximum likelihood on the ground-truth GNSS measurement errors. The classical adaptive filters require no offline training, and their covariance estimates are clipped relative to the receiver-reported covariance to prevent divergence. 


\begin{table}[t]
\centering
\caption{GNSS--INS mean horizontal RMSE improvement (\%) over the nominal KF-GINS filter (\(1.85\)\,m).}
\label{tab:gnss_regimes}
\setlength{\tabcolsep}{3pt}
\footnotesize
\begin{tabular*}{0.8\linewidth}{@{\extracolsep{\fill}}lcrrr@{}}
\toprule
Nominal-law provider & \(\beta\) & DR headroom & Without DR & With DR \\
\midrule
Receiver          & 4.1 & $+0.05$ & $0.0$   & $4.2$ \\
Learned adapter   & 2.7 & $+0.05$ & $16.1$  & $15.7$ \\
Sage--Husa        & 5.2 & $+2.65$ & $-38.6$ & $3.0$ \\
VB-AKF            & 4.0 & $+1.61$ & $-24.0$ & $7.1$ \\
\bottomrule
\end{tabular*}
\end{table}

We evaluate horizontal error on the six outdoor sequences, which are also used to train the learned adapter; the results therefore illustrate mechanism behavior, not out-of-sample generalization. In~\Cref{tab:gnss_regimes}, the nominal KF-GINS RMSE is \(1.85\)\,m; DR-only, adapter-only, and WRAP improve it by \(4.2\%\), \(16.1\%\), and \(15.7\%\).
With the receiver-reported covariance, \(\beta=4.1\) indicates a bias-dominant regime, with approximately \(94\%\) of the nominal filter's squared error associated with the window means.\footnote{Within window \(i\), \(\frac{1}{n_i}\sum_k\|e_{i,k}\|^2=\|\bar e_i\|^2+\Tr\Cov(e_i)\); hence \(\beta^2/(1+\beta^2)\) is the fraction of squared error associated with the window mean.} Meanwhile, $T=0.31$ and a DR-headroom proxy of $+0.05$ indicate little under-dispersion in this diagnostic, consistent with mean adaptation providing most of the point-accuracy gain and WRAP remaining close to adapter-only.

Consistency statistics support this interpretation.\footnote{
For each sequence, NIS and position NEES are averaged over the 3D GNSS updates, and medians of these sequence-level means are reported across the six outdoor sequences; the ideal mean is $3$. NEES uses the post-update position covariance and
post-processed reference. All variants accept the same GNSS updates.} The nominal filter has a median sequence-level mean normalized innovation squared (NIS) of \(0.13\), versus an ideal of \(3.0\), but a median sequence-level mean normalized estimation error squared (NEES) of \(22.7\). Thus, the innovations are much smaller than \(\hat S_t\) predicts, while the true position error exceeds that implied by \(\Xpost\), consistent with measurement bias being absorbed into the state. Adapter-only and WRAP nevertheless lower NEES to \(13.8\) and \(11.6\), while DR-only reaches \(4.0\), closest to the ideal. No variant is fully consistent, suggesting residual mean-model error that the mean-preserving DR step cannot remove.

The same behavior limits innovation-driven adaptation: an absorbed bias can leave innovations apparently well behaved while trajectory error grows. Sage--Husa and VB-AKF therefore reduce the measurement standard deviation to about one tenth of the receiver-reported value, degrading RMSE by \(38.6\%\) and \(24.0\%\). The DR covariance step raises both above the nominal baseline, yielding \(3.0\%\) and \(7.1\%\) improvements. Its effect is small for the receiver and learned adapter, which leave little DR headroom, and largest for the over-tightened classical estimates. This in-sample study illustrates WRAP's intended division of labor: mean adaptation addresses systematic bias, while covariance robustification limits sensitivity to residual covariance misspecification, most visibly for the over-tightened classical estimates.

\subsection{Runtime Evaluation}
\label{subsec:runtime_experiment}

We measure runtime on an NVIDIA Jetson Orin Nano Super Developer Kit with a six-core Arm Cortex-A78AE CPU and a 1024-core Ampere GPU. The adaptive module runs on the GPU, while the C++ Frank--Wolfe solver runs on one CPU thread. All reported times are median latencies over 100 runs after warm-up, measured with CUDA events for inference and wall-clock timing for the covariance solve.

For the TDOA2 sequences of the UTIL dataset, UWB updates arrive at 55\,Hz after downsampling, leaving approximately 18\,ms per update. The DR covariance problem is solved at every accepted measurement update and requires 0.05\,ms. Adaptive inference requires 21--30\,ms, depending on the backbone, but runs at 1\,Hz, with its latest output retained between refreshes.
For the GNSS--INS experiments on the i2Nav-Robot dataset, GNSS updates arrive at 1\,Hz, with IMU propagation at 200\,Hz. Adaptive inference requires 23--26\,ms, while the DR covariance solve requires 2.92\,ms. Both fit comfortably within the 1\,s update interval.

At the evaluated $1$\,Hz adapter refresh, both components meet the timing budgets of the two localization stacks; the refresh rate is configurable.

\section{Conclusion}
\label{sec:conclusion}
WRAP couples causal effective-noise adaptation with a mean-preserving Wasserstein robust gain inside an unchanged EKF/ESKF stack. UWB--IMU results show complementary observed gains, particularly when post-hoc mean and upper-quantile/scale mismatch co-occur. The in-sample GNSS--INS study is bias-dominant: adaptation supplies most of the accuracy gain, while DR improves consistency and mitigates over-tightened classical covariances.

The UWB trials are held out from adapter training but cross-validated for radii; GNSS is in-sample. Moreover, the minimax interpretation is local, radii depend on noise coordinates, and the baseline gate is not robustified. Future work will study frozen out-of-domain tests, normalized or data-calibrated radii, and online regime selection.


\bibliographystyle{IEEEtran}
\bibliography{ref_updated}

\end{document}